%% file: main.tex
\documentclass[letterpaper, 10 pt, conference]{ieeeconf}
\IEEEoverridecommandlockouts
\usepackage{amsmath,amssymb}
\usepackage{graphicx}
\usepackage{xcolor}
\usepackage{booktabs}
\newif\ifshowannotations
\showannotationsfalse
\definecolor{revisionblue}{RGB}{0,80,180}
\definecolor{huajiangreen}{RGB}{0,128,64}
\ifshowannotations

\else

\fi

\usepackage{cuted}
\usepackage[font=footnotesize,labelsep=period]{caption}  %
\usepackage{hyperref}
\hypersetup{
  hidelinks,
  hypertexnames=false,
  pdfauthor={},              %
  pdfsubject={},
  pdfkeywords={},
  pdfborder={0 0 0},          %
  citebordercolor={0 1 0},    %
}
\makeatletter
\NewCommandCopy{\ICRA@cite}{\cite}
\DeclareRobustCommand{\cite}[1]{{\ifshowannotations\hypersetup{pdfborder={0 0 1}}\fi\ICRA@cite{#1}}}
\makeatother

\title{ForkCast: Predictive Recovery for Vision-Language-Action Policies via Counterfactual Futures}

\author{Chenru Wen$^{1,*}$, Huajian Zeng$^{1,*}$,
Haokai Ding$^{1}$, Jian Kang$^{1}$, Anqing Duan$^{1}$, \\ Dawei Zhou$^{2}$, Dezhen Song$^{1}$, Xingxing Zuo$^{1,\dagger}$%
\thanks{$^{1}$ Mohamed bin Zayed University of Artificial Intelligence (MBZUAI), UAE.}%
\thanks{$^{2}$ Virginia Tech, USA.}%
\thanks{$^{*}$ Chenru Wen and Huajian Zeng contributed equally.}%
\thanks{$^{\dagger}$ Xingxing Zuo is the corresponding author (Email: {\tt\small xingxing.zuo@mbzuai.ac.ae}).}
}
\begin{document}
\maketitle

\thispagestyle{empty}
\pagestyle{empty}

\input{sections/0_abstract}

\input{sections/1_intro}

\input{sections/2_related_work}

\input{sections/3_method}

\input{sections/4_experiment}

\input{sections/5_conclusion}

\bibliographystyle{IEEEtran}
\bibliography{references}

\end{document}

%% file: sections/0_abstract.tex
\begin{abstract}
Vision--language--action (VLA) policies perform manipulation tasks using camera images and language instructions, but object slip or visual occlusion can disrupt execution.
Recovery is typically learned during policy training or delegated to a monitor that triggers a scripted repair after detecting a failure.
We introduce ForkCast, a predictive recovery framework that compares counterfactual futures to determine whether and how to intervene in a frozen VLA policy.
At each decision point, ForkCast evaluates three candidates: executing the policy's nominal action chunk (\textsc{Commit}), adjusting the camera viewpoint (\textsc{Reobserve}), and replanning the arm trajectory (\textsc{Replan}).
A candidate-conditioned world model forecasts the latent outcome of each candidate at a common future horizon, and a shared task-progress head scores the predicted outcomes.
By evaluating the policy's own continuation alongside the repair candidates, a single comparison based on task progress determines both whether and how to intervene, without a failure detector, an intervention threshold, or updates to the VLA backbone.
On five simulated RoboTwin~2.0 tasks with injected occlusion and object slip, ForkCast increases pooled success from $13.3\%$ for the frozen $\pi_{0.5}$ policy to $30.0\%$ and transfers to two additional VLA backbones without updating ForkCast.
After adaptation on real-robot candidate rollouts, ForkCast also improves success over the base policy under both clean and visibility-limited conditions. Code will be released to benefit research community.

\end{abstract}

%% file: sections/1_intro.tex
\section{Introduction}
\label{sec:intro}

Vision--language--action (VLA) policies~\cite{intelligence2025pi_,liu2025rdt,cai2026xiaomi} map camera images and a language instruction to chunks of robot actions~\cite{zhao2023learning}.
However, their performance can degrade under out-of-domain (OOD) conditions such as visual occlusion or object displacement~\cite{fei2025libero}.
Occlusion can hide the target, while object displacement can invalidate an ongoing action chunk.
Changing the camera view or correcting the arm motion may help, but neither intervention is appropriate in every case.
This motivates jointly deciding whether to continue the policy or invoke a repair, and which repair to use. We study this decision for a frozen VLA policy equipped with a set of predefined repair primitives, aiming to select the continuation that best advances the task at each action-chunk boundary.

\begin{figure}[t]
  \centering
\includegraphics[width=\linewidth]{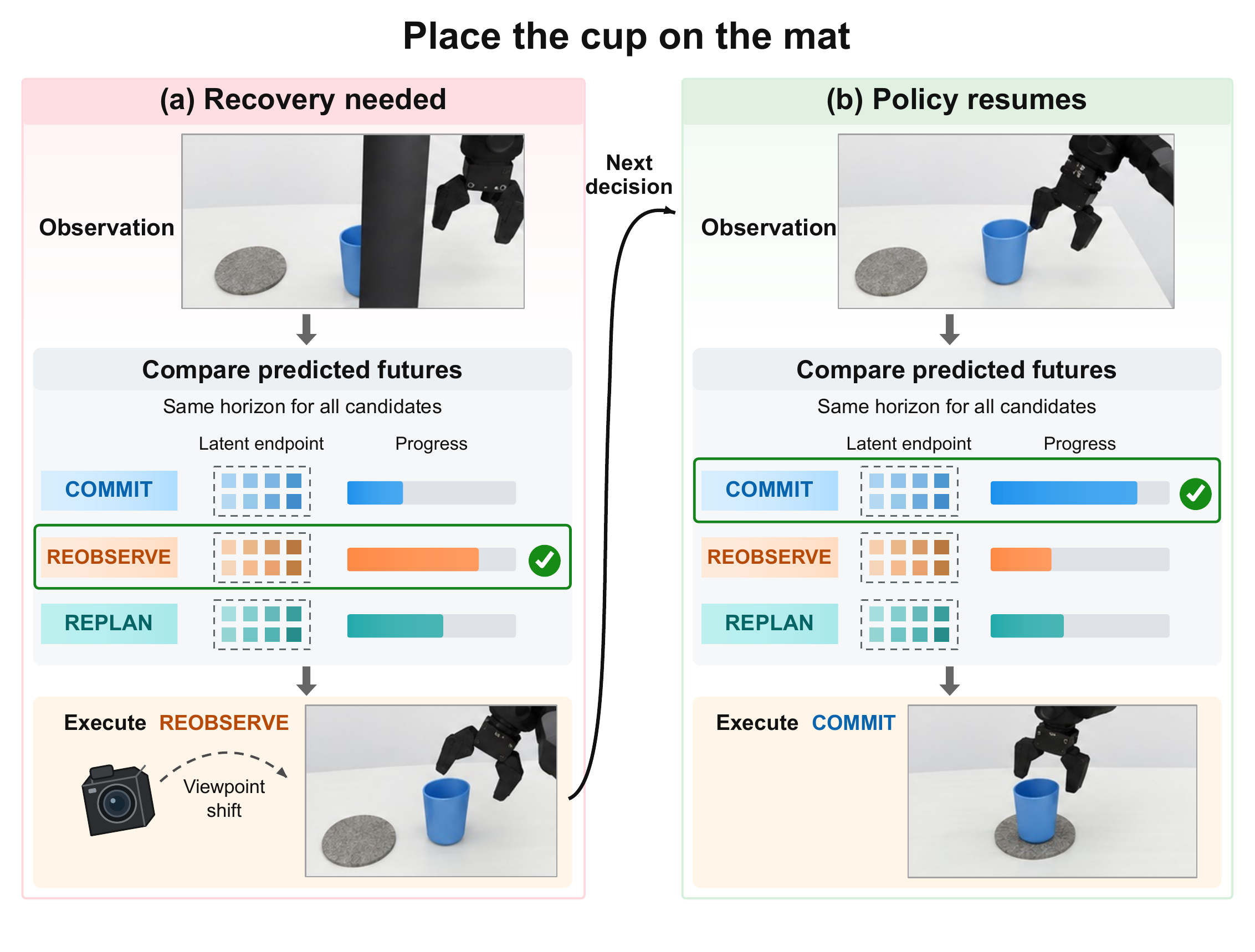}
\captionof{figure}{\textbf{Predictive recovery with ForkCast.}
At a common prediction horizon, ForkCast ranks \textsc{Commit} and available \textsc{Reobserve} and \textsc{Replan} candidates.
(a) Candidate-view acquisition and policy re-query precede scoring. \textsc{Reobserve} ranks highest; ForkCast adopts its selected viewpoint and executes the corresponding cached action chunk.
(b) At the next decision, \textsc{Commit} ranks highest and the frozen VLA policy resumes.
Grids show predicted endpoint latents; bars show relative progress.
Green outlines mark selected candidates.
Views and scores are schematic.}
\label{fig:teaser}
\end{figure}

Existing approaches learn recovery from demonstrations collected under injected disturbances~\cite{laskey2017dart,oh2023bayesian}, train dedicated recovery policies~\cite{vats2025recoverychaining}, or learn camera-control behaviors for active perception~\cite{chuang2025active,liu2026sapave}.
For frozen policies, monitor-routed recovery uses failure signals to trigger repairs~\cite{gu2026safe,xu2025can,zhang2026foresight,zhou2026patch,agia2024unpacking}.
However, a failure signal alone does not establish whether a repair would lead to a better outcome than continuing the policy, nor does it determine which repair would be most effective. This calls for comparing counterfactual futures: how the task would progress under continued policy execution or each available repair.

Action-conditioned world models offer a way to make this comparison by predicting the outcomes of candidate action sequences~\cite{assran2025v,zhou2411dino}. Recent methods use latent predictions to rank VLA-generated chunks alongside predefined motion primitives~\cite{cui2026dreamsteer}, or to score candidate chunks with calibrated progress heads~\cite{zhong2026calibrated}.
Recovery involves two types of repairs: a camera repair changes the observation supplied to the frozen policy and re-queries it for a new action chunk, whereas an arm repair replaces the nominal motion with a planned correction. Predicting these outcomes requires accounting for both the planned arm motion and the camera pose from which the future scene will be well observed. These continuations can also differ in duration, so evaluating each at its own endpoint would compare task progress at different future times.
The technical challenge is therefore to represent both arm and camera motion and compare the resulting continuations over a common prediction horizon.

In this work, we propose \textbf{ForkCast}, a predictive recovery framework for frozen VLA policies (Fig.~\ref{fig:teaser}).
We encode each candidate's planned arm and camera motion over a common prediction horizon, allowing the same predictor to process nominal execution and both repair types. Conditioned on these encodings and the head-camera history, a latent world model adapted from V-JEPA 2-AC~\cite{assran2025v} predicts each candidate’s counterfactual future in the form of an endpoint head-camera latent. A shared progress head maps these predicted latents to task-progress scores.
We train the predictor and progress head on branched rollouts using encoded future observations and scene-based progress labels, encouraging consistent scoring across viewpoints. ForkCast executes the highest-scoring candidate among \textsc{Commit} and the available repairs, jointly deciding whether and how to intervene without a failure detector or intervention threshold.

We evaluate in closed-loop simulation on five RoboTwin~2.0~\cite{chen2025robotwin} tasks with a fine-tuned and then frozen $\pi_{0.5}$~\cite{intelligence2025pi_}, under an injected occlusion that hides the target object from the nominal view and an injected slip that displaces it early in the episode.
Compared with fixed repair strategies built from the same primitives, ForkCast raises pooled success under the two disturbances from $13.3\%$ for the frozen policy and $21.7\%$ for the best fixed strategy to $30.0\%$, and it is best or tied in every condition while no fixed strategy leads under both disturbances.
The model weights of ForkCast, trained on $\pi_{0.5}$ branches, are reused with RDT~\cite{liu2025rdt} and XR-0~\cite{cai2026xiaomi} without further updates, improving pooled success rates on both backbones. 
After adaptation on real-robot candidate rollouts, ForkCast also improves success over the base policy in the real world.

In summary, our contributions are:
\begin{itemize}\setlength{\itemsep}{0pt}\setlength{\parskip}{0pt}
  \item We formulate runtime recovery for a frozen VLA policy as selection among nominal execution, viewpoint-conditioned policy continuations, and planned arm repairs, compared by predicted task progress at a common horizon.
  \item We develop a shared prediction-and-scoring mechanism for heterogeneous recovery candidates. Conditioned on planned arm motion and camera offsets, an adapted latent world model predicts candidate endpoint latents at a common horizon, and a shared progress head maps these latents to task-progress scores.

  \item  We demonstrate that ForkCast raises the success rate of the frozen $\pi_{0.5}$ policy across five simulated RoboTwin~2.0 tasks with injected occlusion and object slip, and transfers to two additional VLA backbones, RDT and XR-0, without updating ForkCast.
\end{itemize}

%% file: sections/2_related_work.tex
\section{Related Work}
\label{sec:related}

\noindent\textbf{Learned Recovery and Active Perception.}
Existing methods train recovery behaviors from disturbance-augmented expert demonstrations~\cite{laskey2017dart,oh2023bayesian}, learn hierarchical recovery policies~\cite{vats2025recoverychaining}, learn camera-control behaviors~\cite{chuang2025active,liu2026sapave}, or optimize the viewpoint at test time before executing the policy~\cite{wang2026observer}.
Image-based methods compensate for viewpoint shifts~\cite{heo2026anycamvla} or occlusion~\cite{li2026libero} while keeping the camera fixed.
ForkCast uses scripted arm and camera repairs and learns to compare their predicted outcomes.

\noindent\textbf{Failure Monitoring.}
Failure monitors derive signals from learned policy-feature scores~\cite{gu2026safe}, conformal out-of-distribution scores~\cite{xu2025can}, or action-conditioned world-model latents~\cite{zhang2026foresight}.
Other systems combine detectors specialized for individual failure modes~\cite{agia2024unpacking}, while chunk-level monitors pause execution before routing to recovery~\cite{zhou2026patch}.
These methods separate detection from repair selection.
ForkCast compares predicted progress across nominal execution and available repairs, using no separate failure detector or intervention threshold.

\noindent\textbf{Predictive Action Selection.}
Action-conditioned world models support planning by predicting candidate outcomes~\cite{assran2025v,zhou2411dino}.
Candidate action chunks can be ranked using learned values or verifiers~\cite{nakamoto2024steering,kwok2025robomonkey,xie2026look}.
Closely related methods evaluate imagined outcomes with vision-language models~\cite{wu2025foresight}, refine diffusion-policy proposals~\cite{qi2026inference}, or score latent rollouts with calibrated progress heads~\cite{zhong2026calibrated}.
DreamSteer~\cite{cui2026dreamsteer} evaluates VLA-generated action chunks and predefined motion primitives through predicted outcomes and value scores. ForkCast additionally evaluates camera-assisted policy continuations, for which changing the viewpoint also changes the policy-generated action chunk. The ForkCast predictor jointly conditions on arm motion and camera offsets, while a shared progress head learns from executed candidate branches. We compare predicted progress over a common horizon.

%% file: sections/3_method.tex
\section{Method}
\label{sec:method}

ForkCast provides runtime recovery for a frozen vision--language--action (VLA) policy by comparing counterfactual futures: the predicted outcomes of alternative candidates from the current decision point.
It evaluates nominal execution and available camera and arm repairs by their predicted task progress.

\begin{figure*}[t]
  \centering
  \includegraphics[width=\textwidth]{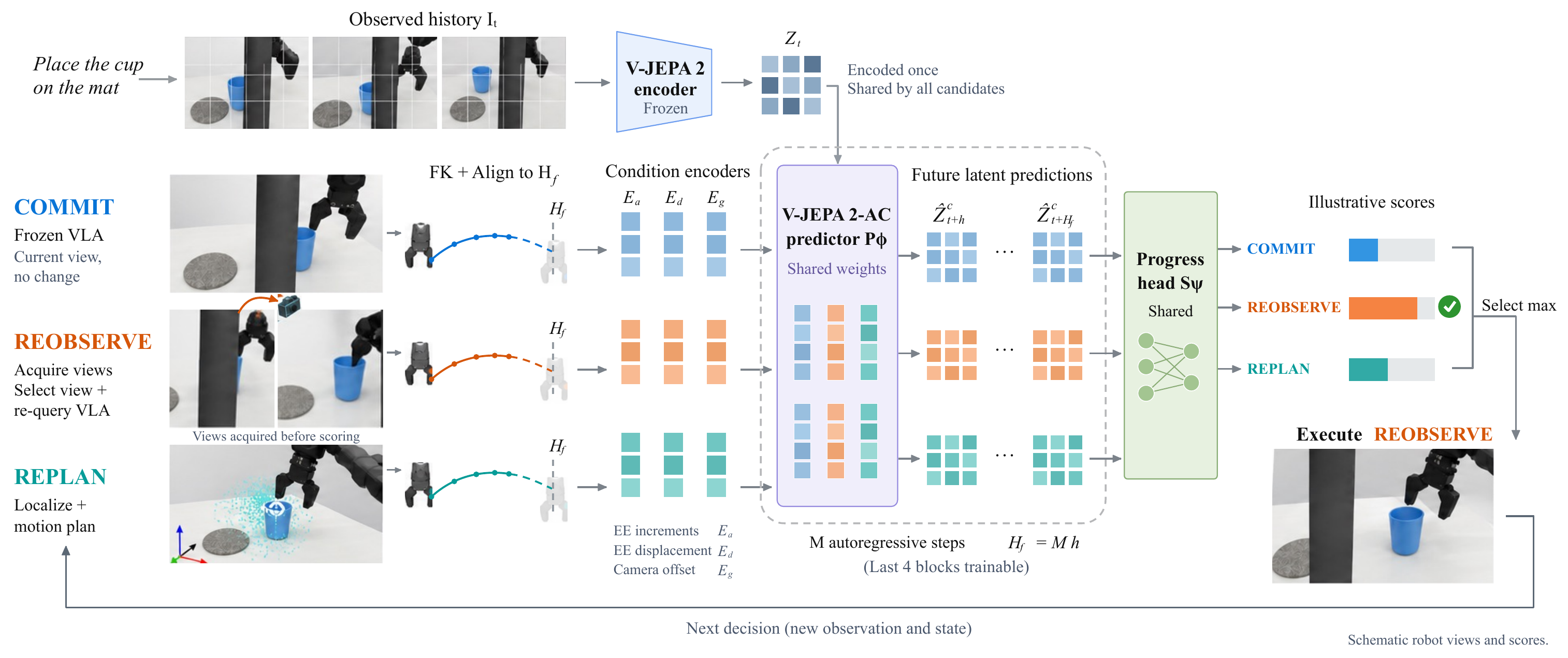}
  \caption{\textbf{Overview of ForkCast.}
\textsc{Commit} retains the nominal plan, \textsc{Reobserve} acquires views and re-queries the frozen policy before scoring, and \textsc{Replan} generates an arm correction.
A frozen visual encoder encodes history once for all candidates.
Candidate arm and camera motion, aligned to a common horizon $H_f$, condition a shared V-JEPA~2-AC predictor~\cite{assran2025v}.
It forecasts $M$ milestones, spaced $h$ control steps apart ($H_f=Mh$).
A shared progress head scores endpoint latents, and ForkCast executes the highest-scoring candidate.
Future grids represent undecoded latents; robot views and scores are schematic.}
  \label{fig:method-overview}
  \vspace{-15pt}
\end{figure*}

As illustrated in Fig.~\ref{fig:method-overview}, ForkCast runs in three stages.
First, candidate generators assemble a candidate set including the frozen policy's nominal action chunk and two types of scripted repairs (Sec.~\ref{sec:method:candidates}).
Second, a candidate-conditioned world model~\cite{assran2025v} predicts each candidate's future head-camera latent, which a learned shared progress head will score (Sec.~\ref{sec:method:prediction}).
Finally, ForkCast executes the highest-scoring candidate after candidate generation and evaluation (Sec.~\ref{sec:method:select}).

\subsection{Problem definition and notation}
\label{sec:problem}

ForkCast augments a frozen VLA policy $\pi_\theta$ with predictive recovery.
At control step $t$, the policy receives an observation $\mathbf{o}_t$, which may include images from multiple cameras, a robot state $\mathbf{s}_t$ containing the arm joint configuration and gripper aperture, and an instruction $\ell$. It predicts an action chunk $\mathbf{A}_t=\pi_\theta(\mathbf{o}_t,\mathbf{s}_t,\ell)$. The nominal policy executes the first $H_a$ control steps before predicting the next action chunk.
We call each action-chunk boundary a \emph{decision point} and denote its arm-control-step index by $t$. 
The visual input of ForkCast is a head-camera history $\mathbf{I}_t=(\mathbf{X}_{\tau_1},\ldots,\mathbf{X}_{\tau_{n_{\mathrm{hist}}}})$ of $n_{\mathrm{hist}}$ RGB frames sampled from the steps leading up to $t$, with $\tau_{n_{\mathrm{hist}}}=t$ and $\mathbf{X}_\tau$ denoting the RGB frame at step $\tau$, while the candidate generators at the first stage additionally read $(\mathbf{o}_t,\mathbf{s}_t,\ell)$.
The history ends before viewpoint probing, and the head-camera pose at that point defines the nominal reference viewpoint.

At the decision point $t$, it assembles the candidate set
\begin{equation}
\begin{aligned}
  \mathcal C_t &= \{\textsc{Commit}\}\cup\mathcal R_t,\\
  \mathcal R_t &\subseteq \{\textsc{Reobserve},\textsc{Replan}\},
\end{aligned}
\label{eq:problem:candidate-set}
\end{equation}
where $\mathcal R_t$ contains only repairs that can be constructed at $t$ (Sec.~\ref{sec:method:candidates}). 
Every candidate $c\in\mathcal C_t$ carries a raw arm trajectory $\mathbf{A}_t^c$ in joint space and, if it requires camera motion, a raw camera trajectory $\mathbf{G}_t^c$.

A frozen visual encoder $E_{\mathrm{VJ}}$ encodes the head-camera history $\mathbf{I}_t$. A learned predictor $P_\phi$ forecasts each candidate's endpoint latent, and a shared progress head $S_\psi$ estimates task progress:
\begin{equation}
\begin{aligned}
  \mathcal Z_t &= E_{\mathrm{VJ}}(\mathbf{I}_t),\\
  \widehat{\mathbf{Z}}_{t+H_f}^{\,c}
    &= P_\phi^{(M)}(\mathcal Z_t,\mathcal U_t^c),\\
  \widehat{\rho}_{t+H_f}^{\,c}
    &= S_\psi(\widehat{\mathbf{Z}}_{t+H_f}^{\,c})\in\mathbb{R}.
\end{aligned}
\label{eq:problem:prediction-score}
\end{equation}
Here, $\mathcal Z_t$ is the encoded observation history, and $P_\phi^{(M)}$ denotes $M$ autoregressive applications of $P_\phi$ using the per-milestone conditioning tokens encoded from $\mathcal U_t^c$ (Sec.~\ref{sec:method:prediction}). 
The prediction horizon begins after candidate-view acquisition and spans $H_f=Mh>H_a$ arm-control steps.
The duration of preliminary viewpoint probing is not included in this horizon.
For candidate $c$, $\widehat{\mathbf{Z}}_{t+H_f}^{\,c}$ represents its predicted endpoint in head-camera latent space, and $\widehat{\rho}_{t+H_f}^{\,c}$ estimates task progress at that endpoint.
The candidate selected for execution at $t$ is
\begin{equation}
  \widehat c_t
  = \arg\max_{c\in\mathcal C_t}\;\widehat\rho^{\,c}_{t+H_f}.
  \label{eq:problem:selection}
\end{equation}

\subsection{Recovery candidates}
\label{sec:method:candidates}

The repair candidates address two complementary recovery needs.
The arm-side need arises when the policy acts on an inappropriate object location, as after a missed grasp or a displaced object~\cite{vats2025recoverychaining}, motivating a planned arm correction.
The camera-side need arises when a task-relevant object is hidden behind an occluder or outside the camera's field of view~\cite{chuang2025active}, motivating a change of camera viewpoint followed by a policy re-query.
The two repairs share a common detection and segmentation visual frontend.

The visual frontend applies an open-vocabulary detector~\cite{minderer2023scaling}, prompted with the target object's name, and a segmenter~\cite{kirillov2023segment} to each image. During candidate construction, the head camera moves to acquire candidate views before prediction and scoring. Probing precedes every selection, including \textsc{Commit}. Metric depth predictions~\cite{yang2024depth}, together with the camera intrinsics and extrinsics during this motion, are used to back-project masked pixels from multiple images into a common 3D coordinate frame, forming a point cloud of the target object. The point cloud centroid, \(\mathbf{p}_t \in \mathbb{R}^{3}\), is computed by averaging all points associated with the object and serves as a target estimate of its position.

\noindent\textbf{\textsc{Commit}.}
The nominal candidate reuses the chunk of arm actions $\mathbf{A}_t$ predicted from the frozen VLA policy and uses the nominal reference viewpoint, restored after probing.

\noindent\textbf{\textsc{Reobserve}.}
The \textsc{Reobserve} candidate moves the head camera to a better viewpoint.
It draws on a library \(\mathcal V\) of \(J\) camera poses \(\mathbf{v}_j\), \(j=1,\ldots,J\), each defined by a lateral offset and a rotation about the yaw axis relative to the nominal reference viewpoint.
The camera-repair generator scores each entry in \(\mathcal V\) by applying the visual frontend to the corresponding image and counting the pixels associated with the target object.
The shortlist \(\mathcal S_t\) retains the indices of entries with a sufficient number of target-object pixels.
The off-center cost \(\operatorname{off}(\mathbf{v}_j)\) is then used to rank the shortlisted views by measuring the distance between the projected target position and the image center.

The accepted head-camera pose minimizes the off-center cost over the shortlist:
\begin{equation}
\hat{\jmath}_t
= \arg\min_{j \in \mathcal{S}_t}
  \operatorname{off}(\mathbf{v}_j),
\qquad
\widehat{\mathbf{v}}_t
= \mathbf{v}_{\hat{\jmath}_t}.
  \label{eq:method:perception-candidate}
\end{equation}
If no entry in $\mathcal V$ provides an adequate view of the target, the generator returns no candidate.
Re-querying the frozen VLA policy from the accepted head-camera viewpoint yields the raw arm trajectory $\mathbf{A}_t^{\textsc{Reobserve}}$, while the raw camera trajectory $\mathbf{G}_t^{\textsc{Reobserve}}$ holds the accepted camera pose throughout the candidate continuation.
Both trajectories are used for subsequent candidate-conditioned endpoint prediction (see Sec.~\ref{sec:method:prediction}).
The generated action chunk is cached and reused if \textsc{Reobserve} is selected.

\noindent\textbf{\textsc{Replan}.}
The replan generator attempts to re-grasp the target object at its current location.
It uses the target position estimate $\mathbf{p}_t$ produced by the visual frontend.
An off-the-shelf motion planner computes a collision-free path from the current robot state to a position near the target object, $\mathbf{p}_t+\mathbf{\Delta}_{\mathrm{tcp}}$, where $\mathbf{\Delta}_{\mathrm{tcp}}\in\mathbb{R}^{3}$ denotes the embodiment-specific fixed flange-to-tool-center displacement.
The planned path preserves the end-effector orientation at the decision point and serves as the raw arm trajectory $\mathbf{A}_t^{\textsc{Replan}}$.
Its endpoint is chosen so that the gripper can enclose the target after \textsc{Replan}.
The replan generator returns no candidate if $\mathbf{p}_t$ is unavailable or planning fails.

\subsection{Counterfactual future prediction and scoring}
\label{sec:method:prediction}
At the second stage of ForkCast, given the shared head-camera history and each candidate's planned arm and camera motion, we predict its counterfactual future as a head-camera latent at the common horizon $t+H_f$.
A progress head shared across candidates and tasks maps each predicted latent to a task-progress score $\widehat{\rho}_{t+H_f}^{c}$.

We initialize the latent predictor from V-JEPA~2-AC, the action-conditioned variant of V-JEPA~2~\cite{assran2025v}. In ForkCast, the predictor is conditioned on planned end-effector increments (from the raw arm trajectory), cumulative end-effector displacement, and camera offsets.

\noindent\textbf{Candidate encoding.}
Forward kinematics maps each raw joint-space trajectory $\mathbf{A}_t^c$ to end-effector poses, from which motion increments are computed.
The candidate representation is $\mathcal U_t^c=(\widetilde{\mathbf A}_t^c,\widetilde{\mathbf D}_t^c,\Delta\mathbf G_t^c)$, comprising per-step end-effector motion increments without gripper commands, cumulative end-effector displacement from the decision state, and camera-pose offsets from the nominal reference viewpoint, respectively. 
For \textsc{Reobserve}, $\Delta\mathbf G_t^c$ repeats the accepted-view offset at every forecast step. \textsc{Commit} and \textsc{Replan} use zero camera offsets after restoring the nominal reference viewpoint.
The operator $\mathrm{Align}_{H_f}$ truncates plans longer than $H_f$ and pads shorter plans by holding their final arm and camera poses.
Branches execute the aligned plan open-loop for $H_f$ steps without re-querying the policy.
The conditioning encoders $E_a$, $E_d$, and $E_g$ produce the corresponding milestone embeddings $\alpha_m^c$, $\delta_m^c$, and $\chi_m^c$, respectively.

Each prediction step spans $h$ control steps. We divide the common horizon of $H_f$ control steps into $M=H_f/h$ segments of $h$ steps each, with a milestone at the end of each segment. Milestones are predicted autoregressively, with each milestone $m$ conditioned on the embeddings of the three conditioning channels, ${\boldsymbol{\alpha}_m^c}$, ${\boldsymbol{\delta}_m^c}$, and ${\boldsymbol{\chi}_m^c}$.

\noindent\textbf{Prediction and scoring.}
The frozen V-JEPA~2 ViT-g encoder $E_{\mathrm{VJ}}$ maps the head-camera history $\mathbf{I}_t$ once per decision point to $\mathcal Z_t=(\mathbf{Z}_1,\ldots,\mathbf{Z}_{T_0})$, comprising $T_0$ latent steps with $N_z$ spatial tokens each, shared by every candidate.
Starting from $\mathcal H_{t,0}^c=\mathcal Z_t$, the predictor $P_\phi$ uses the latent history and the conditioning embeddings through milestone $m$,
\begin{align}
  \widehat{\mathbf{Z}}_{t+mh}^{\,c}
  &= P_\phi\!\bigl(\mathcal H_{t,m-1}^c,
    \boldsymbol{\alpha}_{1:m}^c,
    \boldsymbol{\delta}_{1:m}^c, \boldsymbol{\chi}_{1:m}^c\bigr),
  \label{eq:method:milestone-rollout}\\
  \mathcal H_{t,m}^c
  &= \left(\mathcal H_{t,m-1}^c, \widehat{\mathbf{Z}}_{t+mh}^{\,c}\right),
  \qquad m=1,\ldots,M.
  \label{eq:method:milestone-history}
\end{align}
The subscript $1{:}m$ denotes the corresponding token sequence through milestone $m$.
The progress head $S_\psi$ averages the $N_z$ tokens of a predicted grid and maps them through LayerNorm and a two-layer MLP to $\widehat\rho_{t+mh}^{\,c}=S_\psi(\widehat{\mathbf{Z}}_{t+mh}^{\,c})$.

\noindent\textbf{Training.}
We train $P_\phi$ and $S_\psi$ jointly on rollouts collected by running each available candidate at a decision point.
We refer to this branch collection procedure as \emph{mining}.

The frozen encoder maps each branch's head-camera frame at a milestone to a target latent $\mathbf{Z}_{t+mh}^{c,\mathrm{gt}}$. Progress labels $\rho_{t+mh}^c\in[0,1]$ at selected milestones are computed from simulator ground truth or sensor-derived geometry for real-robot branches.
The progress label is a normalized geometric proxy based on the task-specific distance and shared reference distance described in Sec.~\ref{sec:exp:setup}. For a fixed task, scene state, and normalization reference, this target is independent of camera pose, which encourages consistent progress estimates across viewpoints.

The loss function combines a latent term and a progress term,
\begin{align}
  \mathcal L &= \lambda_Z\mathcal L_Z + \lambda_\rho\mathcal L_\rho,
  \label{eq:method:training-objective}\\
  \mathcal L_\star &= \frac{1}{\max\{1,W_\star\}}\sum_{(t,c)\in\mathcal B}\sum_{m=1}^{M}\omega_m\,\mu^{\star,c}_{t,m}\,e_\star^{\,t,c,m},
  \label{eq:method:training-terms}
\end{align}
with $\star\in\{Z,\rho\}$ and scalar weights $\lambda_Z,\lambda_\rho>0$, where $\mathcal B$ is a batch of candidate branches, $e_Z^{\,t,c,m}$ compares $\widehat{\mathbf{Z}}_{t+mh}^{\,c}$ against $\mathbf{Z}_{t+mh}^{c,\mathrm{gt}}$, and $e_\rho^{\,t,c,m}$ compares $S_\psi(\widehat{\mathbf{Z}}_{t+mh}^{\,c})$ against $\rho_{t+mh}^c$, both under a smooth-$L_1$ loss.
The weights $\omega_m=\exp\!\big(\beta(m-M)\big)$ with $\beta>0$ rise geometrically toward the endpoint and reach one there, emphasizing predictions near the horizon used for selection. The binary masks $\mu^{\star,c}_{t,m}$ select the targets a branch actually recorded, separately for the latent and progress losses. The sum of valid weights, $W_\star=\sum_{(t,c)\in\mathcal B}\sum_{m=1}^{M}\omega_m\,\mu^{\star,c}_{t,m}$, normalizes each term over the entries it sums, with the denominator bounded below by one.
Training is teacher-forced, filling the latent history of Eq.~\ref{eq:method:milestone-history} with target latents, whereas inference rolls out autoregressively.
Progress labels are used only here, and the predictor and progress head read no privileged state at deployment.
\subsection{Selection and execution}
\label{sec:method:select}

\noindent\textbf{Selection.}
ForkCast executes the candidate selected by the argmax in Eq.~\ref{eq:problem:selection}, using only the endpoint score at the final milestone $m=M$ and favoring \textsc{Commit} when it is tied for the maximum.
Selection compares candidates from the same decision state and does not require progress scores to be calibrated across different states or tasks.

\noindent\textbf{Execution.}
\textsc{Commit} executes the first $H_a$ steps of the nominal action chunk without repair.
The camera-pose update for \textsc{Reobserve} does not count toward arm-control steps.
Once \textsc{Reobserve} is selected, we adopt the accepted viewpoint and execute the first $H_a$ steps of the action chunk generated during candidate construction and used for outcome prediction and progress scoring.
The policy is not queried again at execution time.
\textsc{Replan} executes the planned approach and closes the active gripper, then returns control to the frozen policy using the resulting observation and robot state.
Under this convention, \textsc{Replan} is the only candidate that consumes additional arm-control steps beyond those in the policy's action chunk.

%% file: sections/4_experiment.tex
\section{Experiments}
\label{sec:exp}

\subsection{Experimental setup}
\label{sec:exp:setup}

\noindent\textbf{Benchmark and policy.}
We evaluate on RoboTwin~2.0~\cite{chen2025robotwin}, a SAPIEN-based dual-arm tabletop benchmark, in its clean configuration without domain randomization, on five tasks: \emph{Place A-to-B left}, \emph{Place A-to-B right}, \emph{Place cup}, \emph{Stack two blocks}, and \emph{Click bell}.
The base policy is $\pi_{0.5}$~\cite{intelligence2025pi_}, fine-tuned per task with LoRA on $50$ demonstrations and then frozen.

\noindent\textbf{Disturbances.}
Each disturbed episode injects one of two disturbances, one per recovery need of Sec.~\ref{sec:method:candidates}.
\emph{Occlusion} places a thin pillar of half-width $1.8$\,cm on the line of sight from the head camera to the target, which hides the target from the nominal view but not from the lateral views in the camera library.
\emph{Slip} displaces the target by $4$\,cm in a seeded random direction on the table plane at control step $40$.
We call a disturbance type a \emph{condition}, and \emph{clean} denotes episodes with no injected disturbance.

\noindent\textbf{Baselines.}
We compare ForkCast with nominal policy execution, fixed repair strategies, and alternative candidate-selection methods.
\textsc{Commit} executes the frozen policy without repair.
The fixed strategy baselines use predefined recovery schedules with observation-conditioned repair primitives.
Random selects an available candidate randomly.
The Direct progress regressor (Direct) concatenates pooled observation-history and initial-scene features from the same frozen encoder with the proprioceptive state and the candidate plan, and regresses progress with a two-layer MLP (128 GELU units) trained on the same branches and targets as
ForkCast under an MSE loss.
It selects the
highest-scoring available candidate.
The oracle is a privileged reference: at each decision point it executes every candidate from the same simulator state under the mining convention and selects the highest realized progress at $t+H_f$ (ties favor \textsc{Commit}). It bounds selection quality under the proxy, not the attainable success rate.

\noindent\textbf{Protocol.}
All methods are evaluated on the same seeds, disjoint from the mining seeds, using 6 episodes per task and condition across five tasks. This gives 30 episodes per method per condition and 60 episodes for \emph{pooled-OOD}, which combines slip and occlusion. An episode ends when the benchmark success condition is met or after 240 control steps.
Random, the Direct progress regressor, and ForkCast use the same candidate types, observation access, and decision frequency. 

\noindent\textbf{Implementation details.}
We use $H_a{=}32$ and $H_f{=}64$ control steps with $M{=}8$ milestones; ForkCast first decides at step $64$.
The camera-view library for \textsc{Reobserve} comprises $J{=}21$ predefined discrete poses that combine lateral offsets of up to $\pm8$ cm with pan angles of up to $\pm24^\circ$. The library expands to $35$ poses when no target-object position estimate is available.
The visual frontend uses OWLv2~\cite{minderer2023scaling}, SAM~\cite{kirillov2023segment}, and metric Depth Anything~V2~\cite{yang2024depth}.
The progress label is $\rho=\operatorname{clip}(1-d/d_{\mathrm{ref}},0,1)$, where $d$ is the object-to-goal distance (end-effector-to-bell for \emph{Click bell}) and $d_{\mathrm{ref}}$ is the maximum first-milestone distance across branches from the same decision state. This reference is shared by all candidates and milestones from that state and is used only for label construction.
The frozen encoder maps $n_{\mathrm{hist}}{=}16$ frames (stride~$2$) to $T_0{=}8$ latent steps of $N_z{=}256$ tokens; milestone $m$'s conditioning triple occupies latent position $T_0{+}m{-}1$.
We initialize the predictor from V-JEPA~2-AC~\cite{assran2025v} and train its last $4$ blocks, the three conditioning encoders, and the progress head for $30$ epochs using AdamW (batch size~$3$, learning rate $10^{-4}$), with $\beta{=}0.35$, $\lambda_Z{=}\lambda_\rho{=}1$, and progress labels at $m\in\{1,2,4,8\}$.

\subsection{Comparison with baselines}
\label{sec:exp:main}

\begin{table}[t]
\centering
\small
\setlength{\tabcolsep}{3pt}
\begin{tabular}{@{}lcccc@{}}
\toprule
Method & clean & slip & occ & pooled-OOD \\
\midrule
\textsc{Commit}
  & 53.3 & 10.0 & 16.7 & 13.3 \\
Fixed \textsc{Reobserve}
  & 36.7 & 10.0 & 23.3 & 16.7 \\
Fixed \textsc{Replan}
  & 70.0 & \textbf{33.3} & 10.0 & 21.7 \\
\midrule
Random
  & 53.3 & 16.7 & 16.7 & 16.7 \\
\shortstack[l]{Direct progress regressor}
  & 56.7 & 23.3 & 16.7 & 20.0 \\
  ForkCast w/o camera cond. & 70.0 & 26.7 & 16.7 & 21.7 \\
ForkCast (ours)
  & \textbf{76.7} & \textbf{33.3} & \textbf{26.7} & \textbf{30.0} \\
\midrule
Oracle$^\dagger$
  & 83.3 & 36.7 & 26.7 & 31.7 \\
\bottomrule
\end{tabular}
\caption{\textbf{Recovery and candidate-selection comparison.}
Closed-loop success rates (\%) across five RoboTwin~2.0
tasks using the frozen $\pi_{0.5}$ policy.
\emph{Clean} has no injected disturbance; \emph{slip} and
\emph{occ} denote object displacement and visual occlusion.
\emph{Pooled-OOD} averages the two disturbed conditions.
ForkCast w/o camera cond. is retrained without the camera-offset channel.
$^\dagger$The oracle uses privileged outcome information
and is reported separately from deployable methods.
The 95\% Wilson confidence intervals (CIs) for pooled-OOD ($n=60$) are:
ForkCast $[19.9, 42.5]$, Oracle $[21.3, 44.2]$, Fixed REOBSERVE $[9.3, 28.0]$, ForkCast w/o camera $[13.1, 33.6]$,
Fixed REPLAN $[13.1, 33.6]$, Direct $[11.8, 31.8]$,
Random $[9.3, 28.0]$, and COMMIT $[6.9, 24.2]$.}
\label{tab:main}
\end{table}

\begin{figure}[t]
  \centering
  \includegraphics[width=\linewidth]{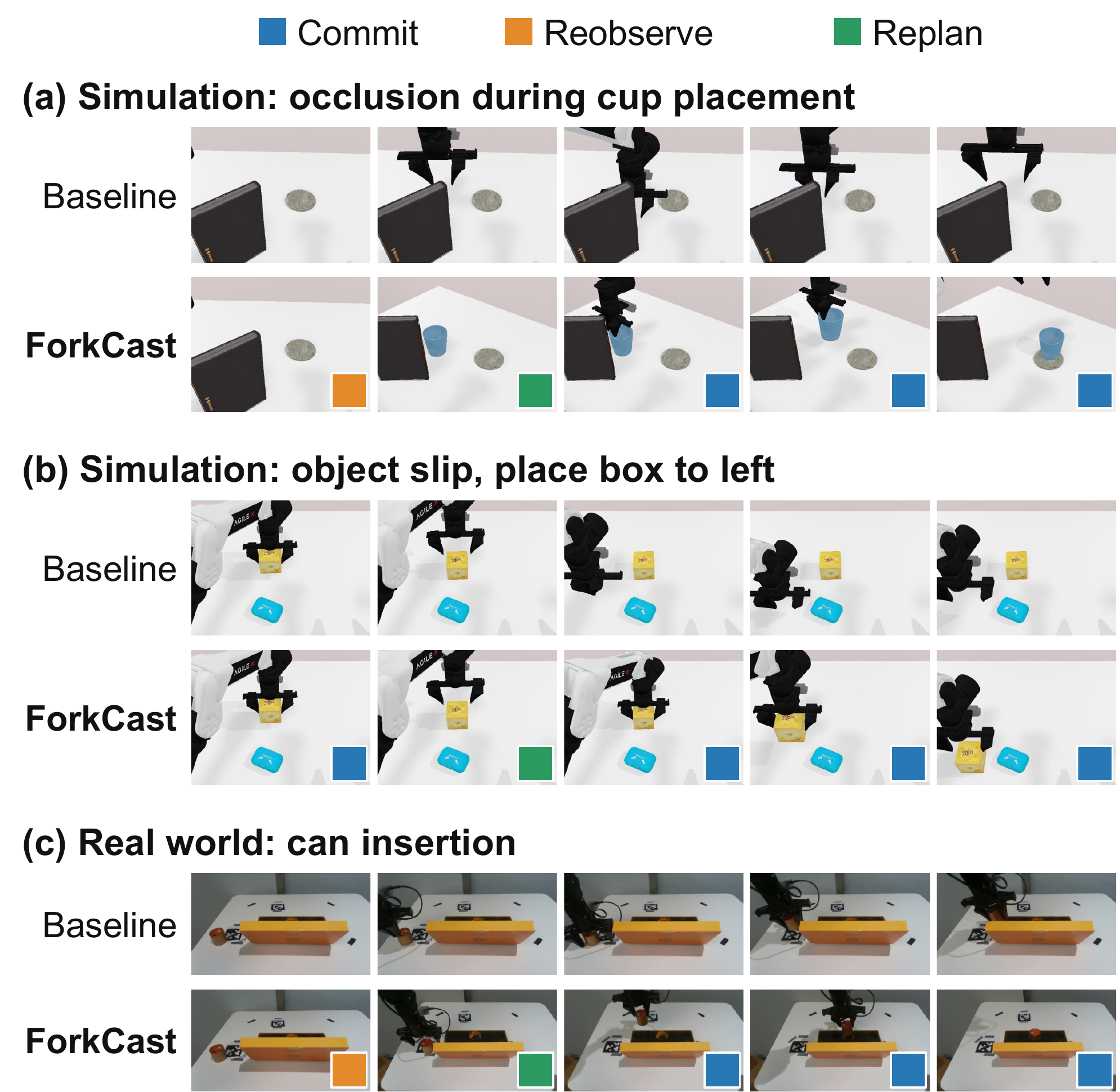}
  \caption{\textbf{Qualitative recovery examples.}
  Head-camera sequences compare the base policy (upper row) with
  ForkCast (lower row) in (a) simulated cup placement under occlusion,
  (b) simulated box placement after object slip, and
  (c) real-world can insertion.
  Frames progress from left to right.
  ForkCast uses \textsc{Reobserve} followed by \textsc{Replan} in (a,c),
  and \textsc{Replan} in (b), before resuming \textsc{Commit}.
  Blue, orange, and green markers denote \textsc{Commit},
  \textsc{Reobserve}, and \textsc{Replan}, respectively.}
  \label{fig:analysis}
\end{figure}

Among fixed strategies, \textsc{Replan} performs best on clean episodes and under slip, while \textsc{Reobserve} performs best under occlusion (Tab.~\ref{tab:main}).
ForkCast reaches $30.0\%$ pooled-OOD success versus $21.7\%$ for the best fixed strategy and $13.3\%$ for the base policy, and outperforms Random and Direct in every condition.
The privileged oracle reaches $31.7\%$, 1.7 points above ForkCast, and matches it under occlusion.
The gaps under slip and clean (3.3 and 6.7 points) reflect residual selection errors.

\subsection{Transfer across policy backbones}
\label{sec:exp:system}

\begin{table}[t]
\centering
\small
\setlength{\tabcolsep}{5pt}
\begin{tabular}{@{}lcccc@{}}
\toprule
Method & clean & slip & occ & pooled-OOD \\
\midrule
XR-0~\cite{cai2026xiaomi}               & 16.7 & 0.0  & 0.0  & 0.0  \\
XR-0 + ForkCast                         & \textbf{56.7} & \textbf{30.0} & \textbf{13.3} & \textbf{21.7} \\
\midrule
RDT~\cite{liu2025rdt}                   & 10.0 & 3.3  & 6.7  & 5.0  \\
RDT + ForkCast                          & \textbf{33.3} & \textbf{33.3} & \textbf{13.3} & \textbf{23.3} \\
\midrule
$\pi_{0.5}$~\cite{intelligence2025pi_}  & 53.3 & 10.0 & 16.7 & 13.3 \\
$\pi_{0.5}$ + ForkCast                  & \textbf{76.7} & \textbf{33.3} & \textbf{26.7} & \textbf{30.0} \\
\bottomrule
\end{tabular}
\caption{\textbf{Cross-backbone transfer in simulation.}
Success rates (\%) on five RoboTwin~2.0 tasks with three frozen policy backbones.
ForkCast trained on $\pi_{0.5}$ branches is applied to RDT and XR-0 without updates.
\emph{Pooled-OOD} averages slip and occlusion.}
\label{tab:system}
\vspace{-10pt}
\end{table}
We apply ForkCast trained on $\pi_{0.5}$ branches to RDT~\cite{liu2025rdt} and XR-0~\cite{cai2026xiaomi} without updating ForkCast (Tab.~\ref{tab:system}).
Each backbone is separately fine-tuned on the same $50$ demonstrations per task.
Repair generators remain unchanged; \textsc{Commit} and \textsc{Reobserve} chunks come from the evaluated backbone.
ForkCast improves every backbone in all conditions, raising pooled-OOD success from $0.0\%$ to $21.7\%$ on XR-0 and from $5.0\%$ to $23.3\%$ on RDT.

\subsection{Ablation study}
\label{sec:exp:ablation}

\begin{table}[t]
\centering
\small
\setlength{\tabcolsep}{4pt}
\begin{tabular}{@{}lcccc@{}}
\toprule
Variant & Sens. & Eff.\ cos & Rank $\rho_s$ & All three \\
\midrule
Full model            & $2.60\times$ & $+.025$ & $+.268$ & yes \\
Replica (retrained)   & $4.08\times$ & $+.018$ & $+.161$ & yes \\
\midrule
\multicolumn{5}{@{}l}{\emph{(a) Conditioning channels}} \\
\textsc{NoCond}       & $\mathbf{0.79\times}$ & $\mathbf{+.001}$ & $+.216$ & no \\
\textsc{NoState}      & $2.55\times$ & $+.015$ & $\mathbf{+.157}$ & no \\
\textsc{NoCamera}     & $\mathbf{1.19\times}$ & $\mathbf{+.001}$ & $+.185$ & no \\
\midrule
\multicolumn{5}{@{}l}{\emph{(b) Training losses}} \\
\textsc{NoL}$_Z$      & $82\times$ & $\mathbf{+.002}$ & $+.184$ & no \\
\textsc{NoL}$_\rho$   & $\mathbf{0.82\times}$ & $+.012$ & $+.239$ & no \\
\midrule
\multicolumn{5}{@{}l}{\emph{(c) Candidate-selection quality}} \\
\multicolumn{2}{@{}l}{Method}
  & $N_{\mathrm{diff}}$ & Hit (\%) & Lost progress \\
\midrule
\multicolumn{2}{@{}l}{ForkCast}  & 16 & 75 & 0.046 \\
\multicolumn{2}{@{}l}{Random}  & 16 & 37.5 & 0.103 \\
\multicolumn{2}{@{}l}{Direct progress regressor}  & 16 & 25 & 0.092 \\
\bottomrule
\end{tabular}
\caption{\textbf{Component ablations and candidate-selection diagnostics.}
(a,b) Conditioning and loss ablations on $45$ held-out mining states, each variant retrained with the channel or loss removed.
Sens., Eff.~cos, and Rank $\rho_s$ denote sensitivity, latent-effect cosine similarity, and Spearman progress correlation.
Replica is an independently retrained full model.
Bold marks a failed criterion.
(c) Candidate-selection diagnostics from $30$ evaluation episodes per method.
$N_{\mathrm{diff}}=16$ counts common candidate sets with unequal realized progress and is the denominator of the best-set hit rate.}
\label{tab:ablation}
\vspace{-10pt}
\end{table}

We ablate conditioning channels and training losses on $45$ mining states from five held-out seeds (Tab.~\ref{tab:ablation}(a,b)).
Each variant is retrained with the corresponding channel zeroed (or loss weight set to zero) and evaluated with the same zeroing at inference.
We measure (i)~\emph{sensitivity}: predicted-progress spread across candidates relative to that under $\sigma{=}5$\,mm jitter of one candidate's target estimate, with a $2\times$ threshold; (ii)~\emph{effect direction}: cosine similarity between predicted and realized candidate-pair latent differences; and (iii)~\emph{rank direction}: within-state Spearman correlation between predicted and realized progress.
The latter two require $95\%$ bootstrap intervals above zero.
The full model and an independently retrained replica satisfy all three criteria.

\noindent\textbf{Conditioning channels (Tab.~\ref{tab:ablation}(a)).}
Zeroing action and camera channels (\textsc{NoCond}) fails the sensitivity and effect-direction criteria. Ranking still passes with the retained displacement channel, which also encodes the arm plan.
Zeroing displacement (\textsc{NoState}) passes sensitivity and effect direction but fails the rank criterion; its correlation ($+.157$) is close to the replica's ($+.161$), and the intervals do not establish a difference between them.
\textsc{NoCamera} reduces sensitivity and latent-effect alignment, but its rank correlation ($+.185$) lies between the full model ($+.268$) and replica ($+.161$); ranking therefore stays within the range of the two full-model runs.
In closed loop, \textsc{Nocamera} reaches $21.7\%$ pooled-OOD, the level of the best fixed strategy, and $16.7\%$ under occlusion, the level of \textsc{Commit}, against $30.0\%$ and $26.7\%$ for the full model (Tab.~\ref{tab:main}). The drop is largest under occlusion, where \textsc{Reobserve} candidates are compared.

\noindent\textbf{Training losses (Tab.~\ref{tab:ablation}(b)).}
Removing $\mathcal L_Z$ yields $82\times$ sensitivity but an effect cosine of only $+.002$, showing that large predicted changes need not align with realized effects.
Removing $\mathcal L_\rho$ eliminates direct progress supervision and reduces sensitivity to $0.82\times$.

\noindent\textbf{Candidate-selection quality
(Tab.~\ref{tab:ablation}(c)).}
ForkCast, Random, and Direct are evaluated on common candidate sets with recorded outcomes.
For each set $\mathcal C$, $\rho^c$ denotes realized endpoint progress and $\widehat c$ the selected candidate.
Best-set hit measures how often $\rho^{\widehat c}=\max_{c\in\mathcal C}\rho^c$ over the $16$ sets with unequal branch outcomes; any maximizing candidate counts as a hit.
Lost progress is $L=\max_{c\in\mathcal C}\rho^c-\rho^{\widehat c}$, with zero indicating a best-branch selection.
ForkCast achieves $75.0\%$ hit rate and $0.046$ mean lost progress, versus $37.5\%$/$0.103$ for Random and $25.0\%$/$0.092$ for Direct (Tab.~\ref{tab:ablation}(c)).
ForkCast hits 12 of the 16 informative sets, Random 6, and Direct 4. 
In the remaining $14$ sets, all candidates realized equal progress, mostly before the object moved, so the proxy assigned identical scores.
Panels~(a) and~(b) show small effect cosines ($\leq 0.03$) and wide rank intervals. We therefore treat these diagnostics as necessary but insufficient checks and rely on panel~(c).

\subsection{Real-world experiments}
\label{sec:exp:realworld}

\begin{figure*}[t]
    \centering
    \includegraphics[width=\textwidth]{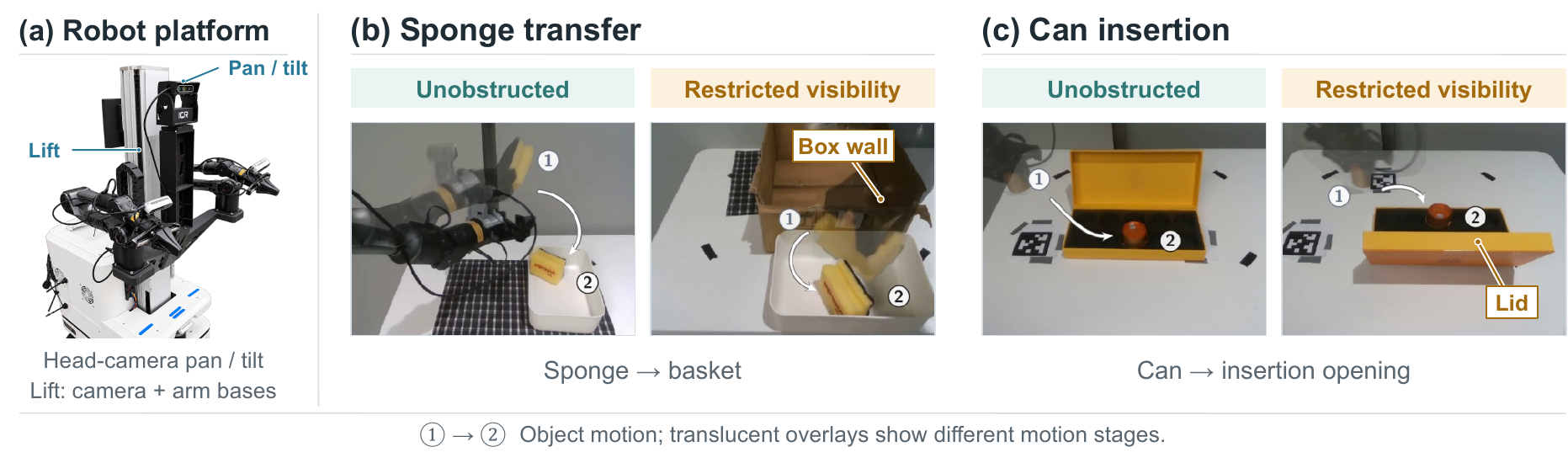}
    \caption{\textbf{Real-world platform and evaluation settings.}
    (a) Dual-arm platform with head-camera pan/tilt and a vertical lift
    carrying the camera and both arm bases.
    (b) Sponge transfer and (c) can insertion, each shown under
    unobstructed (left) and restricted-visibility (right) conditions.
    A box wall restricts visibility of the sponge in (b), while a lid
    restricts visibility of the insertion region in (c).
    Numbered markers and arrows indicate object motion;
    translucent overlays show different motion stages.}
    \label{fig:realworld_setup}
    \vspace{-10pt}
\end{figure*}

\begin{figure}[t]
    \centering
    \includegraphics[width=\columnwidth]{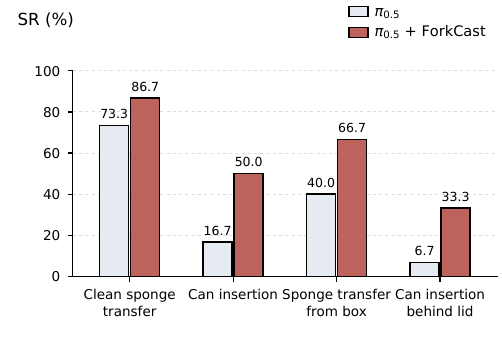}
    \vspace{-2em}
    \caption{\textbf{Real-world success rates with and without ForkCast.}
    Bars report success rates (\%) over $30$ trials per method in each setting.}
    \label{fig:realworld_results}
    \vspace{-15pt}
\end{figure}
\noindent\textbf{Setup and task design.}
Our AgileX PiPER platform has two 6-DoF arms, a rotatable head camera, and a lift carrying the camera and both arm bases; the policy receives head and active-wrist RealSense D435-series images.
We evaluate sponge transfer and can insertion, each with unobstructed or restricted visibility (Fig.~\ref{fig:realworld_setup}).
Unobstructed scenes can still contain execution failures, including can slippage.
A box wall occludes the sponge; a lid restricts visibility during can insertion, combining grasping and visibility challenges.

\noindent\textbf{Training and evaluation.}
We fine-tune $\pi_{0.5}$ on $50$ demonstrations per task, mostly from the canonical view, with a small portion from other viewpoints and coverage of both nominal and raised platform heights.
We adapt simulation-trained ForkCast using $360$ real-robot branches: three candidates from $30$ nominal starting configurations in each of four settings.
Before each branch, objects are manually reset and the recorded preceding actions are replayed to restore the arm configuration.
Object segmentation and RealSense D435i point clouds provide sensor-derived progress estimates using the same normalized-distance formulation as in simulation.
Candidate views are physically acquired before scoring from four options: \emph{canonical}, \emph{left}/\emph{right} (gimbal yaw $\pm10^\circ$), and \emph{raised} (a $10$\,cm lift with arm joints held fixed).
The raise enters $\Delta\mathbf G_t^c$, with $\widetilde{\mathbf D}_t^c$ expressed in the arm-base frame.
The lift takes approximately $5$\,s; either yaw motion takes less than $1$\,s.
If \textsc{Commit} or \textsc{Replan} is selected, the camera and lift return to their pre-probe poses.
We report success over $30$ trials per method per setting for the base policy and ForkCast.

\noindent\textbf{Results.}
ForkCast improves success in all four settings (Fig.~\ref{fig:realworld_results}).
The real-world sequence in Fig.~\ref{fig:analysis} combines \textsc{Reobserve} and \textsc{Replan} before \textsc{Commit} completes insertion, illustrating complementary repairs within one episode.

\subsection{Runtime analysis}
\label{sec:exp:runtime}

We profile $20$ warmed, CUDA-synchronized repetitions on an RTX 6000 Ada, cycling through three recorded histories and evaluating three candidates serially over $M{=}8$ milestones ($64$ control steps).
Mean decision latency is $1.439$\,s, dominated by prediction at $1.355$\,s.
For context, WEAVER~\cite{jain2026weaverbetterfasterlonger} reports $1.020–1.455$ s for four batched candidates at
$H \in \{9, 12, 15\}$ under a different workload and protocol.
Timing covers image loading and preprocessing, CPU--GPU transfers, condition construction, encoding, prediction, and scoring.
It excludes candidate generation, physical viewpoint acquisition or restoration, and execution, and thus measures computation rather than end-to-end robot response.

%% file: sections/5_conclusion.tex
\section{Conclusion}
\label{sec:conclusion}

We presented ForkCast, a predictive recovery framework that enables frozen VLA policies to select among three candidates by forecasting their outcomes: executing the policy's nominal action chunk (\textsc{Commit}), camera repair through \textsc{Reobserve}, and action repair through \textsc{Replan}. By aligning these heterogeneous candidates to a common forecast horizon and scoring their predicted latent outcomes, ForkCast jointly determines whether and how to intervene in policy execution.
Training on counterfactual branches provides supervision for comparing task progress across candidates.
On five RoboTwin 2.0 tasks, ForkCast improves success in clean, occluded, and displaced conditions over the frozen policy and matches or exceeds fixed, random, and direct-regression selection in every condition. Its weights transfer to two further backbones without updates. And after adaptation on real-robot branches, it improves success on a physical dual-arm platform under clean and visibility-limited conditions.

\noindent\textbf{Limitations.}
ForkCast is bounded by the capabilities of its predefined candidates and does not add new manipulation skills to the frozen policy.
In the profiled serial implementation, image-to-selection latency averages $1.439$\,s per decision.
Together with candidate generation and physical viewpoint acquisition, this may limit responsiveness in rapidly changing scenes.
End-to-end deployment latency and the benefits of candidate batching remain to be evaluated.